\documentclass{article}
\usepackage{ijcai25}

\usepackage{times}
\usepackage{soul}
\usepackage{url}
\usepackage[hidelinks]{hyperref}
\usepackage[utf8]{inputenc}
\usepackage[small]{caption}
\usepackage{graphicx}
\usepackage{amsmath}
\usepackage{amsthm}
\usepackage{booktabs}
\usepackage{algorithm}
\usepackage{algorithmic}
\usepackage[switch]{lineno}
\usepackage[table]{xcolor}
\usepackage{multirow}
\usepackage{amssymb}
\usepackage{bbding}

\title{Can Large Models Teach Student Models to Solve Mathematical Problems Like Human Beings? A Reasoning Distillation Method via Multi-LoRA Interaction}

\author{
Xinhe Li$^1$
\and
Jiajun Liu$^{1}$\And
Peng Wang$^{1,2}$\thanks{Corresponding author.}\\
\affiliations
$^1$School of Computer Science and Engineering, Southeast University\\
$^2$Key Laboratory of New Generation Artificial Intelligence Technology and Its \\
Interdisciplinary Applications (Southeast University), Ministry of Education\\
\emails
lixinhe669@gmail.com,
\{jiajliu, pwang\}@seu.edu.cn
}
\newcommand*{\OURMODEL}{LoRID}

\begin{document}

\maketitle

\begin{abstract}
Recent studies have demonstrated that Large Language Models (LLMs) have strong mathematical reasoning abilities but rely on hundreds of billions of parameters.
To tackle the challenge of poor reasoning in Small Language Models (SLMs), existing methods typically leverage LLMs to generate massive amounts of data for cramming training.
In psychology, they are akin to System 1 thinking, which resolves reasoning problems rapidly based on experience and intuition. 
However, human learning also requires System 2 thinking, where knowledge is first acquired and then reinforced through practice.
Inspired by such two distinct modes of thinking, we propose a novel method based on the multi-\textbf{LoR}A \textbf{I}nteraction for mathematical reasoning \textbf{D}istillation (\OURMODEL{}). 
First, we input the question and reasoning of each sample into an LLM to create knowledge-enhanced datasets.
Subsequently, we train a LoRA block on the student model as an Intuitive Reasoner (IR), which directly generates Chain-of-Thoughts for problem-solving.
Then, to imitate System 2 thinking, we train the Knowledge Generator (KG) and Deep Reasoner (DR), respectively.
The former outputs only knowledge after receiving problems, while the latter uses that knowledge to perform reasoning.
Finally, to address the randomness in the generation of IR and DR, we evaluate whether their outputs are consistent, and the inference process needs to be iterated if not.
This step can enhance the mathematical reasoning ability of SLMs through mutual feedback.
Experimental results show that \OURMODEL{} achieves state-of-the-art performance, especially on the GSM8K dataset, where it outperforms the second-best method by 2.3\%, 16.1\%, 2.4\%, 12.3\%, and 1.8\% accuracy across the five base models, respectively.
Meanwhile, we select four strong baselines as System 1, and after integrating them with our method, the reasoning ability of student models is consistently and significantly improved.
The datasets and codes are available at \url{https://github.com/Xinhe-Li/LoRID}.
\end{abstract}

\section{Introduction}

\begin{figure}[t]
    \centering
    \includegraphics[width=0.96\linewidth]{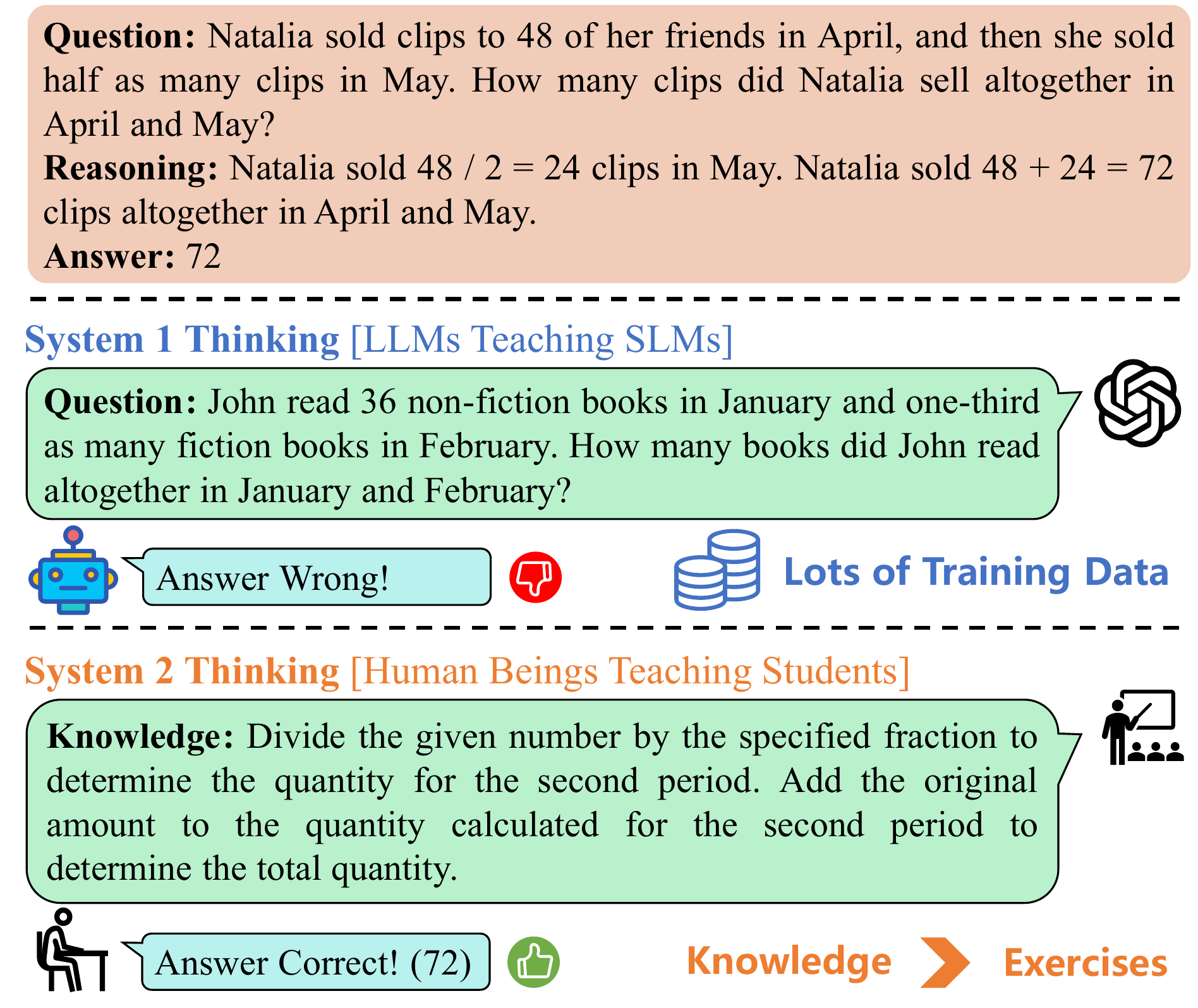}
    \caption{The LLMs teaching SLMs learning pattern vs. Human beings teaching students learning pattern.} 
    \label{fig:introduction}
\end{figure}

Large Language Models (LLMs)~\cite{gpt4,gemini1} have demonstrated superiority in mathematical reasoning with the help of Chain-of-Thought (CoT)~\cite{cot,zero-shot-cot} prompts.
However, although these closed-source models have strong capabilities in a variety of Natural Language Processing tasks, such as semantic understanding~\cite{semantic1,semantic2}, instruction following~\cite{instruction1,instruction2}, and code generation~\cite{codex,code-interpreter}, they rely on hundreds of billions of parameters.
This makes these models undeployable at scale due to overwhelming computational requirements and inference costs~\cite{cot}.
Although Small Language Models (SLMs) have fewer parameters, they face the challenge of poor reasoning ability. 
For example, LLaMA-2-7B~\cite{llama2} and Mistral-7B~\cite{mistral} have only 14.6\% and 15.5\% accuracy on the GSM8K~\cite{gsm8k} dataset after in-context learning~\cite{gpt3}.
Therefore, how to effectively distill the mathematical reasoning ability of teacher models into SLMs is still a non-trivial problem.

To address this issue, existing works~\cite{metamath,mugglemath,wizardmath} mainly use powerful LLMs to perform various data augmentations on CoTs (e.g., Monte Carlo Tree Search~\cite{mcts,mcts2}) and distill reasoning capabilities into the student model through supervised fine-tuning.
Meanwhile, some methods~\cite{mumath-code,mammoth,tora} highlight the synergy between LLMs and external tools (e.g., code interpreter) to reduce computational errors.
They train SLMs with extensive programming language data to develop code-generation capabilities.

However, as shown in Figure~\ref{fig:introduction}, the LLMs teaching SLMs learning pattern is fundamentally different from the human beings teaching students learning pattern.
In psychology, there are two thinking modes: System 1 and System 2~\cite{thinking}. 
The former typically generates quick but error-prone results, while the latter reasons through a slower and deeper thought process.
Inspired by this, on one hand, the data augmentation process of most methods does not explicitly induce the knowledge and capabilities of teacher language models, which contrasts with the way humans transfer knowledge.
Taking the math problem in Figure~\ref{fig:introduction} as an example, they require LLMs to generate several similar questions based on the original question as a training set, instead of imitating teachers to explicitly tell the knowledge to students, which is crucial in the deep thinking of System 2.
On the other hand, the model distillation process does not fully consider the interaction between System 1 and System 2, which is contrary to the way humans acquire knowledge.
Intuition and deep thinking often play different roles in reasoning, and thus their complementarity aids in problem-solving.
Meanwhile, although tool-based methods achieve good performance in tasks involving complex computations, they often promote excessive dependence on external tools~\cite{neurosymbolic} and need to repeatedly send the code generated by student models to a compiler until it executes correctly.

To deal with the above issues, inspired by the human beings teaching and learning pattern, we propose a novel method based on the multi-\textbf{LoR}A~\cite{lora} \textbf{I}nteraction for mathematical reasoning \textbf{D}istillation (\OURMODEL{}).
First, we construct the training sets by prompting a closed-source teacher model (e.g., GPT-4) with zero-shots~\cite{zero-shot} to generate the knowledge required to solve math problems.
Secondly, analogous to System 1, we train a LoRA block on the student model as the Intuitive Reasoner (IR), directly generating Chain-of-Thought, similar to most data augmentation-based methods.
Thirdly, analogous to System 2, we train Knowledge Generator (KG) and Deep Reasoner (DR), respectively. 
These two modules are designed to imitate the processes of students learning knowledge and applying that in practice.
Finally, inspired by the integration of System 1 and System 2 in human learning, if the outputs of IR and DR are inconsistent, the three LoRA blocks mentioned above will continue to iteratively infer until the termination conditions are met.
Through this multi-LoRA interaction on the same student model, they continuously provide feedback to each other, thereby enhancing the overall problem-solving ability in a parameter-efficient manner.

We conduct experiments on the GSM8K~\cite{gsm8k} and MATH~\cite{math} datasets using LLaMA-2-7B~\cite{llama2}, LLaMA-3-8B~\cite{llama3}, Mistral-7B~\cite{mistral}, Qwen2.5-Math-7B~\cite{qwen2.5math}, and DeepSeekMath-7B~\cite{deepseekmath} as our base models.
Experimental results demonstrate that the interaction between System 1 and System 2 significantly enhances the mathematical reasoning abilities of student models.
Especially on the GSM8K dataset, \OURMODEL{} outperforms the second-best method by 2.3\%, 16.1\%, 2.4\%, 12.3\%, and 1.8\% accuracy across the five base models. 
Furthermore, due to the plug-and-play flexibility of LoRA blocks, we select four strong baselines (MuggleMath~\cite{mugglemath}, MuMath~\cite{mumath}, MetaMath~\cite{metamath}, and RFT~\cite{rft}) as System 1, and after integrating our method, the accuracy of student models shows consistent and significant improvement.

The main contributions of this paper are three-fold:
\begin{itemize}
    \item We focus on the mathematical reasoning distillation task and propose a novel method \OURMODEL{}, to the best of our knowledge, which is among the first to draw inspiration from the human beings teaching and learning pattern.
    \item We introduce knowledge during data augmentation and propose multi-LoRA interaction during model distillation, which improves the student's reasoning abilities.
    \item Experimental results show that with the interaction between System 1 and System 2, \OURMODEL{} outperforms previous state-of-the-art approaches and can be easily and effectively integrated into any CoT distillation method.
\end{itemize}

\section{Related Work}

\begin{figure*}[t]
    \centering
    \includegraphics[width=\linewidth]{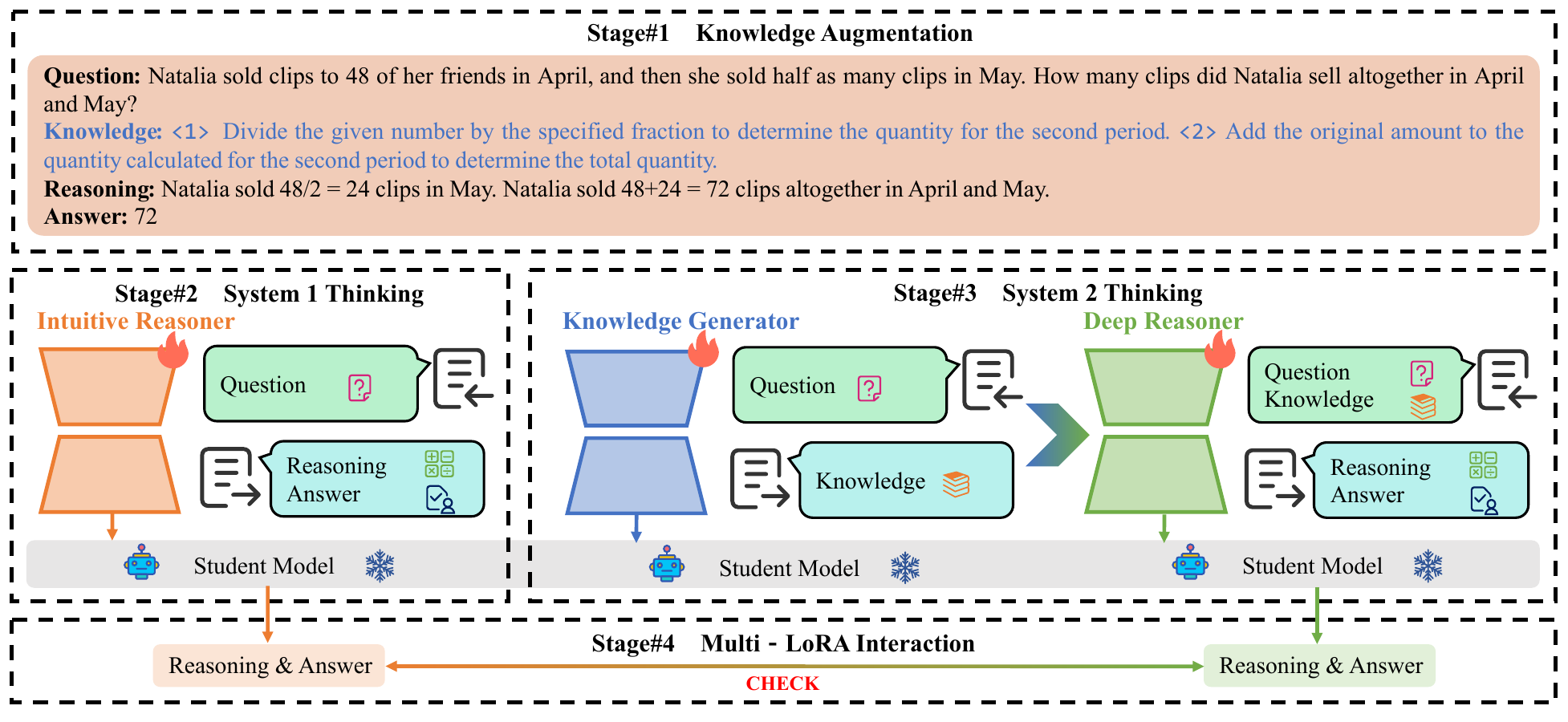}
    \caption{Overview of our proposed \OURMODEL{} framework.}
    \label{fig:framework}
\end{figure*}

Mathematical reasoning tasks like GSM8K~\cite{gsm8k} and MATH~\cite{math} are among the most challenging problems in LLMs.
To solve them, recent works~\cite{cot,zero-shot-cot} show that it is possible to elicit reasoning abilities by prompting LLMs to perform Chain-of-Thought (CoT) reasoning, i.e., generate a series of intermediate steps, but it reduces the accuracy of models with less than 10 billion parameters.
Thus, most current methods~\cite{mugglemath,mathscale} mainly use mainstream closed-source LLMs to generate diverse and high-quality enhanced data.
MuMath~\cite{mumath} and MetaMath~\cite{metamath} bootstrap the questions in both forward and backward reasoning directions.
They require LLMs to produce a large volume of reasoning data, which raises both augmentation and training costs.

Another research trajectory~\cite{mumath-code,mammoth} highlights the synergy between LLMs and external tools.
ToRA~\cite{tora} interleaves Python code blocks and natural language reasoning parts in multiple rounds of the same solution, which provides a more flexible combination of CoT and Program-of-Thought (PoT).
Although using a compiler to output the final answer helps reduce computational errors, it requires the student model to repeatedly generate code until it compiles correctly.
Furthermore, if the SLM is only pre-trained on natural language texts, rather than programming languages, it will be difficult to enable the model to master coding capabilities based solely on supervised fine-tuning.
Therefore, this paper does not consider the use of external tools.

\section{Methodology}

\subsection{Preliminary}
A mathematical reasoning problem can be denoted as $\mathcal{D} = \{(q_i,r_i,a_i)\}_{i=1}^n \subseteq \mathcal{Q} \times \mathcal{R} \times \mathcal{A}$, where each sample includes question $q_i$, reasoning $r_i$, and answer $a_i$.
Our task is to train a student model $f(q_i;\theta)\to [r_i\oplus a_i]$ with parameters $\theta$ to minimize the prediction loss $\mathcal{L}$, which can be formulated as:
\begin{equation}
    \mathcal{L}=\frac{1}{n}\sum_{i=1}^{n}\ell(f(q_i;\theta),[r_i\oplus a_i]).
\end{equation}
where $\ell$ is the cross entropy loss between predicted tokens and target tokens, and $n$ is the amount of data.
Then we compare the answer $a_i$ with $\hat{a}_i$ generated by models to evaluate their mathematical reasoning ability.

\subsection{Framework}
The framework of \OURMODEL{} is shown in Figure~\ref{fig:framework}, which mainly includes four stages: knowledge augmentation, System 1 thinking, System 2 thinking, and multi-LoRA interaction.
In stage 1, we use a closed-source LLM as a teacher to generate knowledge-enhanced mathematical reasoning datasets.
The question and reasoning are provided as prompt inputs to inspire LLMs to output the knowledge for problem-solving.
In stage 2, similar to most other methods, we train a LoRA block to generate a series of reasoning steps (i.e., CoT) for intuitive reasoning.
In Stage 3, we separately train a Knowledge Generator to imitate the process of students acquiring knowledge, and a Deep Reasoner to apply that knowledge to solve mathematical problems.
In stage 4, since the three LoRA blocks mentioned above are trained on the same student model, they can be plug-and-play during inference, allowing them to interact in a parameter-efficient manner.
By comparing the responses from System 1 and System 2, we determine whether further inference is required.
This is similar to how students in human society need to rely not only on intuition but also on deep thinking to reason.

\subsection{Knowledge Augmentation}

The human learning process can be divided into two steps: 
(1) acquiring knowledge to solve a specific type of problem, and
(2) practicing with exercises to flexibly apply that knowledge.
However, the current CoT distillation paradigm~\cite{teaching-slm-to-reason} only generates a large number of data using an LLM and then directly fine-tunes student models on these problems, which deviates from the way humans learn.
Thus, motivated by this, we aim to explicitly extract knowledge from the teacher model.

\begin{figure}[htb]
    \centering
    \includegraphics[width=\linewidth]{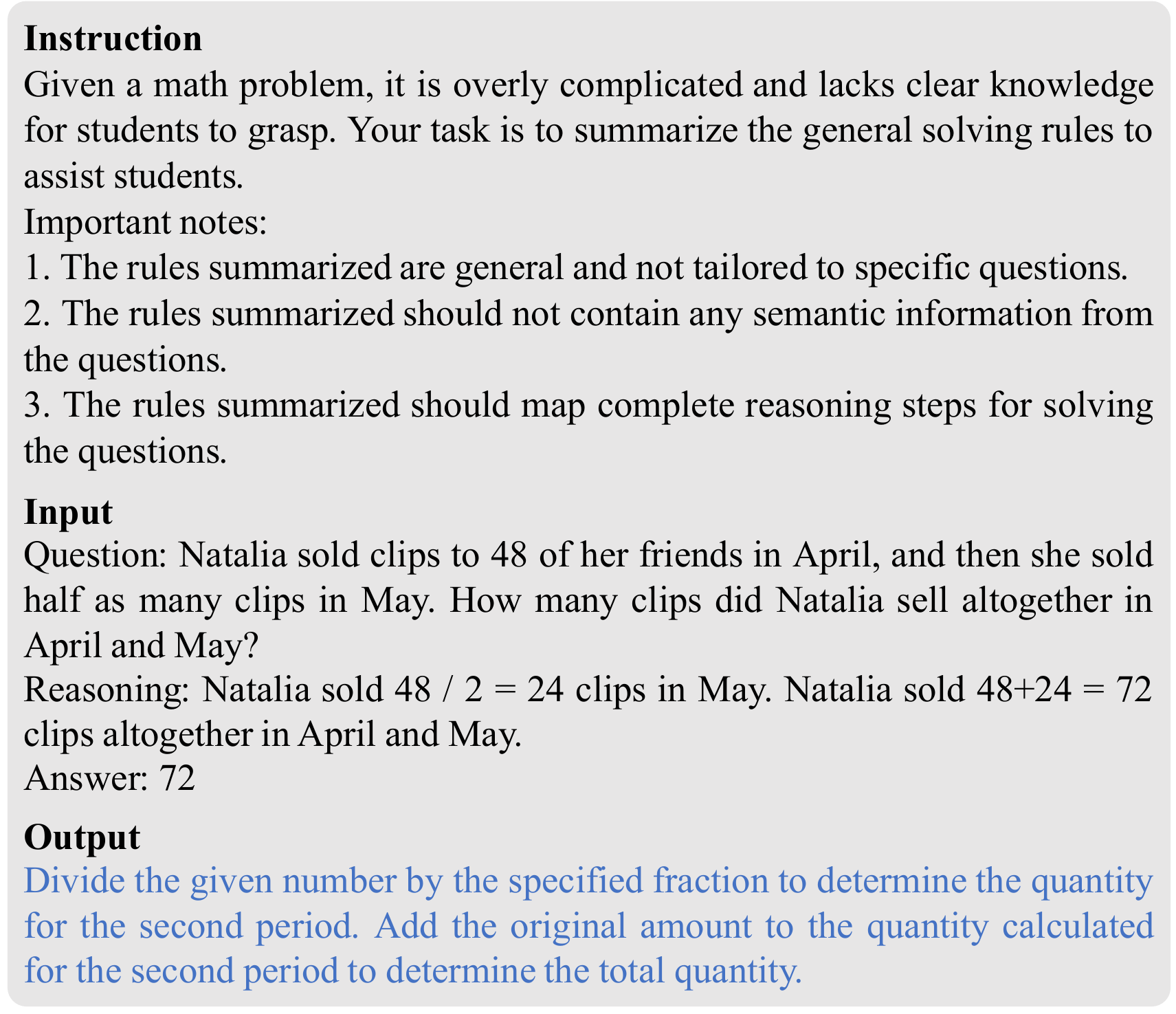}
    \caption{The format of the knowledge generation prompt.}
    \label{fig:knowledge_augmentation}
\end{figure}

Consider a standard sample $d_i$ consisting of a question $q_i$, its correct
reasoning $r_i$ and answer $a_i$.
As shown in Figure~\ref{fig:knowledge_augmentation}, we use zero-shot~\cite{zero-shot} instructions $I$ to prompt a teacher model to generate the general knowledge $k_i$ required to solve this problem.
Since most language models undergo extensive pre-training on raw text, our knowledge representation is also in the form of natural language.
For any dataset $\mathcal{D}$, the entire process can be formulated as follows:
\begin{equation}
    f_{\text{LLM}}(\mathcal{P})\to\mathcal{K}.
\end{equation}
where $\mathcal{P}=\{(I,q_i,r_i,a_i)\}_{i=1}^n$ denotes a prompt set and $\mathcal{K}=\{k_i\}_{i=1}^n$ denotes a knowledge set.

\subsection{System 1 Thinking}
% Learning also requires a lot of practice to consolidate knowledge.
In system 1, similar to other approaches~\cite{mumath,metamath}, we train an Intuitive Reasoner (IR) specifically designed for mathematical reasoning.
The input consists of a question $q_i$, and the output is the concatenation of reasoning $r_i$ and answer $a_i$.
We train a student model $f(q_i;\theta_{W_{\text{IR}}})\to [r_i\oplus a_i]$ to minimize the prediction loss $\mathcal{L}_\text{IR}$:
\begin{equation}
    \mathcal{L}_\text{IR}=\frac{1}{n}\sum_{i=1}^{n}\ell(f(q_i;\theta_{W_{\text{IR}}}),[r_i\oplus a_i])
\end{equation}
\begin{equation}
    W_{\text{IR}} := W_{\text{init}} + A_{\text{IR}}B_{\text{IR}}
\end{equation}
where $W_{\text{init}}\in \mathbb{R}^{d\times k}$ denotes a pre-trained weight matrix of the student model, $A_{\text{IR}}\in\mathbb{R}^{d\times r}$ and $B_{\text{IR}}\in\mathbb{R}^{r\times k}$ are LoRA parameters of Intuitive Reasoner, and the rank $r\ll\min(d,k)$.
In this phase, the student model directly learns the problem-solving skills necessary for later comparison with the output from the Deep Reasoner. 

\subsection{System 2 Thinking}
In System 2, inspired by human learning, a first-grade student can only attempt to answer a fifth-grade math problem based on his existing knowledge. 
However, due to the lack of more advanced knowledge, solving the problem correctly becomes challenging. 
Thus, acquiring additional knowledge is essential for effective problem-solving.

For Knowledge Generator (KG), we take the question $q_i$ as input and the knowledge $k_i$ as output, training a student model $f(q_i; \theta_{W_{\text{KG}}})\to k_i$ to minimize the prediction loss $\mathcal{L}_\text{KG}$:
\begin{equation}
    \mathcal{L}_\text{KG}=\frac{1}{n}\sum_{i=1}^{n}\ell(f(q_i;\theta_{W_{\text{KG}}}),k_i)
\end{equation}
\begin{equation}
    W_{\text{KG}} := W_{\text{init}} + A_{\text{KG}}B_{\text{KG}}
\end{equation}
where $A_{\text{KG}}\in\mathbb{R}^{d\times r}$ and $B_{\text{KG}}\in\mathbb{R}^{r\times k}$ are LoRA parameters of the Knowledge Generator.
During this phase, the student model learns the essential knowledge required for solving problems from the teacher.
Since the semantic complexity of the problem has been simplified, knowledge exhibits less diversity than reasoning, making it easier for students to grasp general rules.
Without explicit knowledge, students would struggle to generalize from a large number of problems and may rely more on rote memorization.

It is widely known that acquiring knowledge enhances problem-solving abilities, but practice is also necessary for students to fully internalize this knowledge. 
For Deep Reasoner (DR), we concatenate the question $q_i$ and knowledge $k_i$ as the input, with reasoning $r_i$ and answer $a_i$ as the output.
The student model $f([q_i \oplus k_i];\theta_{W_\text{DR}})\to[r_i \oplus a_i]$ is trained to minimize the prediction loss $\mathcal{L}_\text{DR}$:
\begin{equation}
    \mathcal{L}_\text{DR}=\frac{1}{n}\sum_{i=1}^{n}\ell(f([q_i \oplus k_i];\theta_{W_\text{DR}}),[r_i \oplus a_i])
\end{equation}
\begin{equation}
    W_{\text{DR}} := W_{\text{init}} + A_{\text{DR}}B_{\text{DR}}
\end{equation}
where $A_{\text{DR}}\in\mathbb{R}^{d\times r}$ and $B_{\text{DR}}\in\mathbb{R}^{r\times k}$ are LoRA parameters of Deep Reasoner.
In the training phase, knowledge is generated by closed-source LLMs, while in the inference phase, it is provided by the Knowledge Generator.

\subsection{Multi-LoRA Interaction}

Just as in student learning, some mathematical problems can be solved using System 1, while others require System 2, which involves first learning the necessary knowledge and then solving the problems.
Inspired by this process, integrating System 1 and System 2 is beneficial for the reasoning of student models.
Since the three LoRA blocks, Intuitive Reasoner, Knowledge Generator, and Deep Reasoner, are fine-tuned on the same model, their plug-and-play advantage facilitates parameter-efficient interactive inference.

In terms of implementation, we store the answers produced by Intuitive Reasoner and Deep Reasoner in $\mathcal{A}_{\text{IR}}$ and $\mathcal{A}_{\text{DR}}$, respectively, during each iteration.
When both sets have the identical answer $\hat{a}_i$, the result is considered final, and inference stops; otherwise, the process continues.
To manage inference costs, we set a threshold $t$ to limit the number of iterations.
Unlike existing methods, we do not require the Intuitive Reasoner or Deep Reasoner to produce highly accurate outputs in a single iteration; rather, we only require the final solution to be correct.
This idea reduces the need for extensive training data and computational time.
Similarly, humans cannot solve problems on the first attempt and typically require multiple trials and errors to find the correct answer.

\section{Experiments}

\subsection{Experimental Setup}

\subsubsection{Datasets}
We use two popular mathematical reasoning benchmarks:
(1) GSM8K~\cite{gsm8k} consists of high-quality grade school math word problems, containing 7,473 training samples and 1,319 test samples;
and (2) MATH~\cite{math} dataset consists of high school competition problems covering seven subjects, and contains 7,500 and 5,000 samples for training and testing, respectively.
Problems in GSM8K require between 2 and 8 steps to get an answer, while MATH is much more challenging.

For each sample in datasets, we call GPT-4o~\cite{gpt4} to generate the knowledge sequence required to solve the problem.
To increase the amount of data, we directly use subsets obtained by MetaMathQA~\cite{metamath} based on answer augmentation and question rephrasing.
Since the two augmentations do not significantly modify the reasoning steps, data from the same original problem can share the knowledge we generate.
Thus, we obtain 7,473 pieces of knowledge for GSM8K and 7,500 pieces of knowledge for MATH.
The statistics of datasets are shown in Table~\ref{tab:datasets}.

% Table generated by Excel2LaTeX from sheet 'datasets'
% Table generated by Excel2LaTeX from sheet 'Datasets'
\begin{table}[t]
  \centering
\setlength{\tabcolsep}{10pt}
  \resizebox{\linewidth}{!}{
    \begin{tabular}{l|ccc}
    \toprule
    Dataset & Training & \#GSM8K & \#MATH \\
    \midrule
    MuggleMath~\cite{mugglemath} & System 1 & 152,589 & 147,787 \\
    MuMath~\cite{mumath} & System 1 & 384,261 & 366,244 \\
    MetaMath~\cite{metamath} & System 1 & 240,000 & 155,000 \\
    RFT~\cite{rft}   & System 1 & 103,638 & - \\
    \midrule
    Ours  & System 2 & 160,000 & 125,000 \\
    \bottomrule
    \end{tabular}%
    }
    \caption{Statistics of datasets for training System 1 and System 2.}
  \label{tab:datasets}%
\end{table}%

\subsubsection{Baselines}
We compare \OURMODEL{} with some strong baselines, which are divided into three groups. 
(1) \textbf{Closed-source models}, we compare GPT-4o, GPT-o1-mini, Claude 3.5 Sonnet, Gemini 1.5-Pro, and DeepSeek-V3 with in-context learning.
(2) \textbf{Open-source models with tools}, we provide 8 baseline methods for comparison: ToRA~\cite{tora}, MAmmoTH~\cite{mammoth}, MathCoder~\cite{mathcoder}, $\text{R}^3$~\cite{r3}, MathGenieLM~\cite{mathgenielm}, MuMath-Code~\cite{mumath-code}, OpenMath~\cite{openmath}, and AlphaMath~\cite{alphamath}, which all require the help of code compilers to output the answer.
(3) \textbf{Open-source models without tools}, we make comparisons with the following state-of-the-art baselines including RFT~\cite{rft}, MetaMath~\cite{metamath}, QDMR~\cite{qdmr}, AutoPRM~\cite{autoprm}, MuMath~\cite{mumath}, MathScale~\cite{mathscale}, $\text{R}^3$, MFT~\cite{mft}, Math-Shepherd~\cite{math-shepherd}, MuggleMath~\cite{mugglemath}, DPO-ST~\cite{dpo-st}, AlphaMath, RefAug~\cite{refaug}, Self-Refine~\cite{self-refine}, and DART-Math~\cite{dart-math}.
Additionally, we conduct experiments on three general models, LLaMA-2-7B, Mistral-7B, and LLaMA-3-8B~\cite{llama3}, as well as two math-specialized models, Qwen2.5-Math-7B~\cite{qwen2.5math} and DeepSeekMath-7B~\cite{deepseekmath}.
However, methods like OVM~\cite{ovm}, which require combining up to 100 outputs to achieve more accurate results, are not included in our comparison.

\subsubsection{Settings}
All experiments are conducted on the 8 $\times$ NVIDIA A100 GPUs.
We set the rank and $\alpha$ of LoRA to 512 and 1024 respectively.
We employ the AdamW~\cite{adamw} optimizer with a cosine learning rate schedule spanning a total of 5 epochs of training. 
The maximum learning rate is set at 5e-5 and there is a 3\% linear warmup.
Considering the diversity of generation, we set the top-p and temperature during inference to 0.90 and 1.50.
The inference iteration threshold $t$ for multi-LoRA interaction is set to 20.

\subsection{Main Results}
We conduct comparative experiments to evaluate the performance of each method in the mathematical reasoning task. 
Table~\ref{tab:main} shows the accuracy results of all models on the GSM8K and MATH datasets.

% Table generated by Excel2LaTeX from sheet 'main'
\begin{table}[!htb]
  \centering
  \setlength{\tabcolsep}{6pt}
  \resizebox{\linewidth}{!}{
    \begin{tabular}{lcccc}
    \toprule
    Method & Base model & \#params & GSM8K & MATH \\
    \midrule
    \multicolumn{5}{c}{\textit{Closed-source models}} \\
    ICL   & GPT-4o & - & 92.9 & 76.6 \\
    ICL   & GPT-o1-mini & - & 94.8 & 90.0 \\
    ICL   & Claude 3.5 Sonnet & -    & 96.4 & 71.1 \\
    ICL   & Gemini 1.5-Pro & - & 91.7  & 58.5 \\
    ICL   & DeepSeek-V3 & 671B & 89.3 & 61.6 \\
    \midrule
    \multicolumn{5}{c}{\textit{Open-source models with tools}} \\
    ToRA  & LLaMA-2 & 7B    & 68.8  & 40.1 \\
    MAmmoTH & LLaMA-2 & 7B    & 53.6  & 31.5 \\
    MathCoder & LLaMA-2 & 7B    & 64.2  & 23.3 \\
    $\text{R}^3$ & LLaMA-2 & 7B    & 68.9  & - \\
    MathGenieLM & LLaMA-2 & 7B    & 71.7  & 33 \\
    MuMath-Code & LLaMA-2 & 7B    & 83.8  & 48.8 \\
    MAmmoTH & Mistral & 7B    & 75.0  & 40.0 \\
    MathGenieLM & Mistral & 7B    & 80.5  & 45.1 \\
    OpenMath & Mistral & 7B    & 80.2  & 44.5 \\
    AlphaMath & DeepSeekMath & 7B    & 84.1  & 66.3 \\
    \midrule
    \multicolumn{5}{c}{\textit{Open-source models without tools}} \\
    ICL   & LLaMA-2 & 7B    & 14.6  & 2.5 \\
    SFT   & LLaMA-2 & 7B    & 41.6  & 7.2 \\
    RFT   & LLaMA-2 & 7B    & 51.2  & - \\
    MetaMath & LLaMA-2 & 7B    & 66.5  & 19.8 \\
    QDMR  & LLaMA-2 & 7B    & 30.4  & - \\
    AutoPRM & LLaMA-2 & 7B    & 70.8  & 23.6 \\
    MuMath & LLaMA-2 & 7B    & \underline{76.2}  & 23.3 \\
    MathScale & LLaMA-2 & 7B    & 66.3  & \textbf{31.1} \\
    $\text{R}^3$ & LLaMA-2 & 7B    & 50.5  & - \\
    MFT   & LLaMA-2 & 7B    & 69.0  & 20.8 \\
    Math-Shepherd & LLaMA-2 & 7B    & 73.2  & 21.6 \\
    MuggleMath & LLaMA-2 & 7B    & 69.8  & 23.1 \\
    DPO-ST & LLaMA-2 & 7B    & 54.7  & - \\
    \rowcolor{blue!20}
    \textbf{Ours} & LLaMA-2 & 7B    & \textbf{78.5} & \underline{25.2} \\
    \midrule
    ICL   & LLaMA-3 & 8B    & 58.4  & 17.0 \\
    SFT   & LLaMA-3 & 8B    & 60.9  & 18.1 \\
    DPO-ST & LLaMA-3 & 8B    & 68.8  & - \\
    AlphaMath & LLaMA-3 & 8B    & \underline{71.8}  & \underline{41.9} \\
    \rowcolor{blue!20}
    \textbf{Ours} & LLaMA-3 & 8B    & \textbf{87.9} & \textbf{44.7} \\
    \midrule
    ICL   & Mistral & 7B    & 15.5  & 10.1 \\
    SFT   & Mistral & 7B    & 50.3  & 13.4 \\
    MetaMath & Mistral & 7B    & 77.7  & 28.2 \\
    MathScale & Mistral & 7B    & 74.8  & \underline{35.2} \\
    MFT   & Mistral & 7B    & 79.5  & 29.0 \\
    Math-Shepherd & Mistral & 7B    & \underline{81.8}  & 33.0 \\
    RefAug & Mistral & 7B    & 78.9  & 30.1 \\
    Self-Refine & Mistral & 7B    & 71.6  & - \\
    \rowcolor{blue!20}
    \textbf{Ours} & Mistral & 7B    & \textbf{84.2} & \textbf{38.7} \\
    \midrule
    ICL   & Qwen2.5-Math & 7B    & 57.7  & \underline{52.1} \\
    SFT   & Qwen2.5-Math & 7B    & \underline{79.4}  & 49.1 \\
    \rowcolor{blue!20}
    \textbf{Ours} & Qwen2.5-Math & 7B    & \textbf{91.7} & \textbf{61.2} \\
    \midrule
    ICL   & DeepSeekMath & 7B    & 65.7  & 33.4 \\
    SFT   & DeepSeekMath & 7B    & 67.2  & 30.9 \\
    DART-Math & DeepSeekMath & 7B    & \underline{88.2}  & \underline{52.9} \\
    \rowcolor{blue!20}
    \textbf{Ours} & DeepSeekMath & 7B    & \textbf{90.0} & \textbf{54.8} \\
    \bottomrule
    \end{tabular}
    }
  \caption{Accuracy results (\%) of the compared methods on GSM8K and MATH datasets (ICL: In-context learning, SFT: Supervised fine-tuning on the training set of GSM8K or MATH). Results of baselines are retrieved from original papers. The \textbf{bold} scores indicate the best results and \underline{underlined} scores indicate the second best results.}
  \label{tab:main}%
\end{table}%
First, compared to the open-source models without tools, \OURMODEL{} outperforms all other baselines across all datasets, except MathScale.
Specifically, on the GSM8K dataset, it achieves accuracy improvements of 2.3\%, 16.1\%, 2.4\%, 12.3\%, and 1.8\% over the second-best methods when deployed on the LLaMA-2-7B, LLaMA-3-8B, Mistral-7B, Qwen2.5-Math-7B, and DeepSeekMath-7B base models respectively.
This demonstrates that our method benefits from the interaction between System 1 and System 2, and is more effective than others based solely on CoT data augmentation.
Furthermore, considering that our method is implemented based on LoRA blocks, we can choose better data augmentation methods to train System 1 and flexibly try different LoRA combinations, so there is still potential for performance improvement.
Although MathScale (31.1\%) has higher accuracy than our method (25.2\%) on the MATH dataset, their training data size is approximately 2 million, much more than we require.

Second, \OURMODEL{} achieves significant performance improvements on different base models, including general models such as LLaMA-3-8B and math-specialized models such as DeepSeekMath-7B.
On the GSM8K dataset, our method outperforms the zero-shot context learning method by 63.9\%, 29.5\%, 68.7\%, 34.0\%, and 24.3\% on the LLaMA-2-7B, LLaMA-3-8B, Mistral-7B, Qwen2.5-Math-7B, and DeepSeekMath-7B base models respectively.
Additionally, compared to closed-source LLMs with hundreds of billions of parameters, the open-source models trained based on our method are already close in mathematical reasoning capabilities (e.g, 91.7\% on Qwen2.5-Math-7B vs. 92.9\% on GPT-4o).
It shows that imitating the way teachers impart knowledge in CoT distillation is effective, and the student model even surpasses the teacher in some capabilities.

Finally, the experimental results demonstrate that \OURMODEL{} has consistent improvements on both the GSM8K dataset, which emphasizes natural language understanding, and the MATH dataset, which focuses on mathematical calculations.
However, taking Mistral-7B as an example, on the GSM8K dataset, the accuracy of our method is 3.7\% higher than the best open-source models with tools, but 6.4\% lower on the MATH dataset.
This indicates that for datasets (e.g., MATH) involving complex calculations, tool-based methods have certain advantages due to leveraging the capabilities of external tools. 
For datasets (e.g., GSM8K) that emphasize knowledge reasoning but involve simple calculations, their performance is inferior to the method we proposed, suggesting that their reasoning abilities remain insufficient.

\subsection{Ablation Results}

We conduct ablation experiments on the GSM8K and MATH datasets, where System 1 is trained on the MuggleMath, MuMath, MetaMath, and RFT augmented datasets, and System 2 is trained on the knowledge-enhanced reasoning dataset we constructed.
As shown in Table~\ref{tab:ablation}, it is noticed that \OURMODEL{} outperforms the methods trained with only CoT (w/o System 2) by 4.4-25.0\% on LLaMA-2-7B and 3.6-22.4\% on Mistral-7B across all datasets.
This indicates that for some problems, students need to first learn the knowledge and then apply it to answer (i.e., System 2).
Additionally, \OURMODEL{} achieves higher accuracy than methods that do not use System 1, which demonstrates that the integration of both systems is necessary. 
The LoRA blocks, trained on the same student model, provide the foundation for implementing this interaction.
Finally, taking the GSM8K dataset as an example, the performance of Mistral-7B in the \OURMODEL{}, System 1, and System 2 is improved by 5.7\%, 7.0\%, and 6.0\% compared with LLaMA-2-7B.
The performance gain brought by the base model itself is consistent across each module of our method.

% Table generated by Excel2LaTeX from sheet 'Ablation'
\begin{table}[tb]
  \centering
  \setlength{\tabcolsep}{5pt}
  \resizebox{\linewidth}{!}{
    \begin{tabular}{cc|cccc}
    \toprule
        \multicolumn{2}{c|}{\multirow{2}[4]{*}{Method}} & \multicolumn{2}{c}{LLaMA-2-7B} & \multicolumn{2}{c}{Mistral-7B} \\
\cmidrule{3-6}    \multicolumn{2}{c|}{} & GSM8K & MATH  & GSM8K & MATH \\
    \midrule
    \multirow{3}[2]{*}{MuggleMath} & LoRID & \textbf{0.785} & \textbf{0.252} & \textbf{0.832} & \textbf{0.387} \\
          & w/o System 1 & 0.597 & 0.148 & 0.667 & 0.217 \\
          & w/o System 2 & 0.741 & 0.201 & 0.789 & 0.351 \\
    \midrule
    \multirow{3}[2]{*}{MuMath} & LoRID & \textbf{0.783} & \textbf{0.231} & \textbf{0.842} & \textbf{0.352} \\
          & w/o System 1 & 0.597 & 0.148 & 0.667 & 0.217 \\
          & w/o System 2 & 0.700 & 0.151 & 0.773 & 0.259 \\
    \midrule
    \multirow{3}[2]{*}{MetaMath} & LoRID & \textbf{0.726} & \textbf{0.203} & \textbf{0.785} & \textbf{0.316} \\
          & w/o System 1 & 0.597 & 0.148 & 0.667 & 0.217 \\
          & w/o System 2 & 0.647 & 0.124 & 0.679 & 0.221 \\
    \midrule
    \multirow{3}[2]{*}{RFT} & LoRID & \textbf{0.682} &    -   & \textbf{0.743} & - \\
          & w/o System 1 & 0.597 &    -   & 0.667 & - \\
          & w/o System 2 & 0.432 &     -  & 0.519 & - \\
    \bottomrule
    \end{tabular}%
    }
    \caption{Ablation results on the LLaMA-2-7B and Mistral-7B base student model. Since RFT has not augmented data for the MATH dataset, there are no related experimental results.}
  \label{tab:ablation}%
\end{table}%

\subsection{Discussions}

\subsubsection{Analysis of Scaling Laws}
We use LLaMA-2-7B and Mistral-7B as our base model to study the scaling laws of \OURMODEL{}.
In the experiment, System 1 is trained with augmented data from MuggleMath, MuMath, MetaMath, and RFT, with data sizes set to 7.5k, 40k, 80k, and 140k, respectively. 
The training data sizes for the Knowledge Generator and Deep Reasoner in System 2 are also consistent with those of System 1.
Table~\ref{tab:scaling_laws} shows that, with the same number of training samples, our approach outperforms those that rely solely on System 1 for reasoning, after incorporating System 2.
Using 40k samples, \OURMODEL{} consistently achieves better results than other methods that use 140k samples.
Additionally, we observe that as the data size increases, the performance of our method shows an upward trend in most cases, but eventually reaches a plateau, which is consistent with the findings of most other works~\cite{mugglemath}.
When the sample sizes are 7.5k, 40k, 80k, and 140k, our method achieves an average accuracy improvement of 11.8\%, 11.0\%, 9.1\%, and 5.6\% compared to the baselines, respectively. 
This suggests that \OURMODEL{} may have greater potential for application in low-resource settings.

% Table generated by Excel2LaTeX from sheet 'Scaling Laws'
\begin{table}[ht]
  \centering
  \setlength{\tabcolsep}{2pt}
  \resizebox{\linewidth}{!}{
    \begin{tabular}{c|cccc|cccc}
    \toprule
    \multirow{2}[4]{*}{Method} & \multicolumn{4}{c|}{LLaMA-2-7B} & \multicolumn{4}{c}{Mistral-7B} \\
\cmidrule{2-9}          & 7.5k  & 40k   & 80k   & 140k  & 7.5k  & 40k   & 80k   & 140k \\
    \midrule
    MuggleMath w/ S-1 & 0.562 & 0.689 & 0.719 & 0.731 & 0.738 & 0.776 & 0.793 & 0.808 \\
    MuggleMath w/ S-1\&2 & \textbf{0.637} & \textbf{0.742} & \textbf{0.762} & \textbf{0.778} & \textbf{0.798} & \textbf{0.832} & \textbf{0.829} & \textbf{0.821} \\
    \midrule
    MuMath w/ S-1 & 0.470 & 0.596 & 0.653 & 0.694 & 0.661 & 0.719 & 0.762 & 0.776 \\
    MuMath w/ S-1\&2 & \textbf{0.600} & \textbf{0.699} & \textbf{0.744} & \textbf{0.763} & \textbf{0.770} & \textbf{0.819} & \textbf{0.810} & \textbf{0.812} \\
    \midrule
    MetaMath w/ S-1 & 0.462 & 0.590 & 0.616 & 0.636 & 0.632 & 0.703 & 0.716 & 0.696 \\
    MetaMath w/ S-1\&2 & \textbf{0.592} & \textbf{0.691} & \textbf{0.699} & \textbf{0.731} & \textbf{0.748} & \textbf{0.795} & \textbf{0.777} & \textbf{0.772} \\
    \midrule
    RFT w/ S-1 & 0.419 & 0.443 & 0.486 &   -    & 0.541 & 0.576 & 0.559 & - \\
    RFT w/ S-1\&2 & \textbf{0.566} & \textbf{0.638} & \textbf{0.663} &   -    & \textbf{0.720} & \textbf{0.757} & \textbf{0.745} & - \\
    \bottomrule
    \end{tabular}%
    }
    \caption{Performance of \OURMODEL{} using different sizes of training data on the GSM8K dataset (S-1: System 1, S-2: System 2). Since the augmented dataset of RFT is less than 140k, there are no relevant experimental results. }
  \label{tab:scaling_laws}%
\end{table}%

\subsubsection{Analysis of Problem Difficulty}
\begin{figure}[tb]
    \centering
    \includegraphics[width=\linewidth]{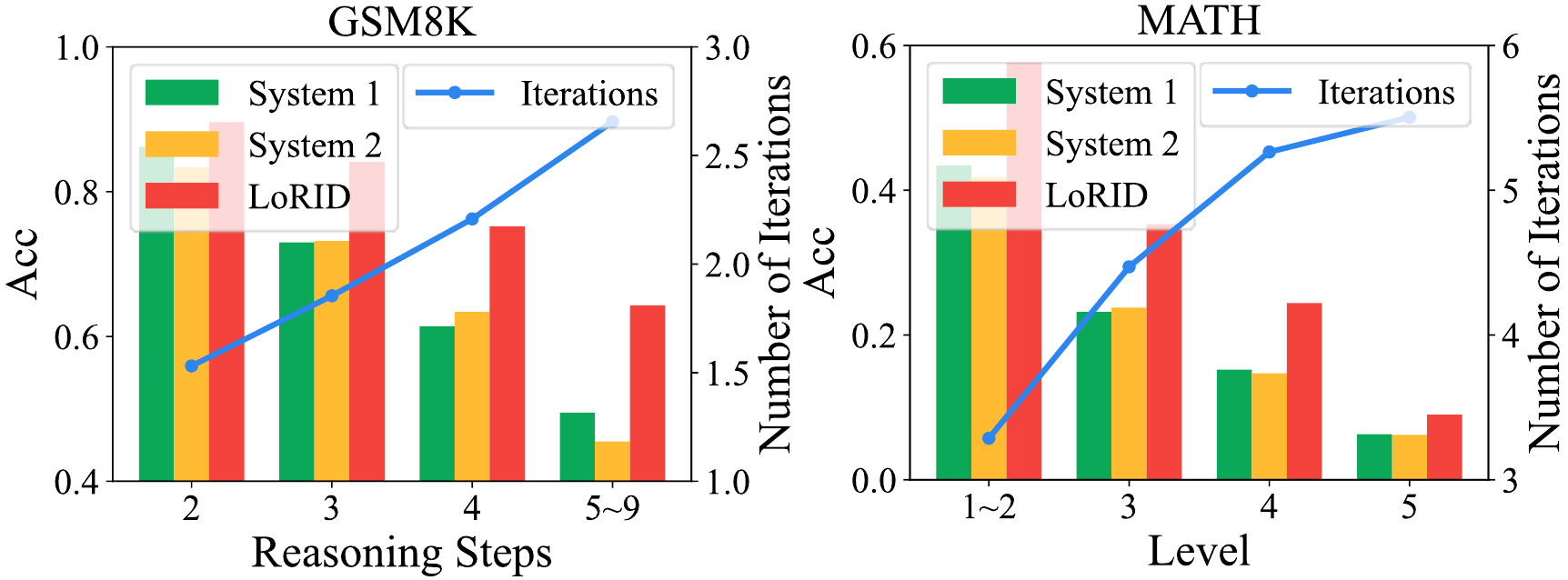}
    \caption{Performance of \OURMODEL{} on different problem difficulties.}
    \label{fig:problem_difficulty}
\end{figure}

We investigate the effectiveness of \OURMODEL{} on problems with varying difficulties, with experiments conducted on Mistral-7B, while System 1 is trained on the MetaMath augmented dataset.
The GSM8K is categorized by the number of reasoning steps, while the MATH has five levels of difficulty, ranging from low to high.
In Figure~\ref{fig:problem_difficulty}, across all levels of problem difficulty, our method improves the reasoning accuracy by an average of 10.6\% and 11.8\% compared to System 1 and System 2, respectively.
This indicates that the integration of two thinking modes enabled by LoRA blocks contributes to the improvement of the student model's mathematical reasoning ability.
Furthermore, we observe that more difficult problems require more iterations.
This aligns with the common sense: students tend to engage in multiple rounds of self-reflection and correction when facing hard problems.

\subsubsection{Analysis of Inference Cost}
We analyze the performance of \OURMODEL{} and Self-consistent CoT (SC-CoT)~\cite{cot-sc}, then further demonstrate our feasibility.
In Table~\ref{tab:inference_cost}, our method achieves an accuracy improvement of 9.2\% and 12.4\% compared to System 1 and System 2, respectively, which perform inference once.
We trade a small increase in inference time for improved accuracy, a concept that has recently been adopted by some large models, such as OpenAI o1~\cite{openaio1}.
Furthermore, \OURMODEL{} even achieves comparable performance to SC-CoT (k=10) on both models, which shows that the interaction between System 1 and System 2 is more efficient in terms of inference overhead than the interaction within a single system alone (e.g., two Intuitive Reasoners).
We speculate that System 1 is better suited for handling problems based on experience and intuition, while System 2 is more suitable for problems grounded in reasoning and logic. The preferences of the two systems for solving problems differ, and this will be explored in future work.

% Table generated by Excel2LaTeX from sheet 'Inference Cost'
\begin{table}[htb]
  \centering
  \setlength{\tabcolsep}{8pt}
  \resizebox{\linewidth}{!}{
    \begin{tabular}{cc|cccc}
    \toprule
    \multicolumn{2}{c|}{\multirow{2}[4]{*}{Method}} & \multicolumn{2}{c}{LLaMA-2-7B} & \multicolumn{2}{c}{Mistral-7B} \\
\cmidrule{3-6}    \multicolumn{2}{c|}{} & Acc $\uparrow$  & \# Iter $\downarrow$ & Acc $\uparrow$  & \# Iter $\downarrow$ \\
    \midrule
    \multirow{3}[2]{*}{System 1} & SC-CoT (k=1) & 0.649 & 1     & 0.679 & 1 \\
          & SC-CoT (k=5) & 0.718 & 5     & 0.757 & 5 \\
          & SC-CoT (k=10) & \textbf{0.739} & 10    & 0.782 & 10 \\
    \midrule
    \multirow{3}[2]{*}{System 2} & SC-CoT (k=1) & 0.597 & 1     & 0.667 & 1 \\
          & SC-CoT (k=5) & 0.667 & 5     & 0.732 & 5 \\
          & SC-CoT (k=10) & 0.704 & 10    & 0.754 & 10 \\
    \midrule
    \multicolumn{2}{c|}{\OURMODEL{}} & 0.727 & 2.3   & \textbf{0.785} & 2.1 \\
    \bottomrule
    \end{tabular}
    }
    \caption{Performance of \OURMODEL{} and Self-consistent CoT on
the GSM8K dataset.}
  \label{tab:inference_cost}%
\end{table}%

\subsubsection{Case Study}
We conduct a case study to verify that System 2 can compensate for the errors caused by intuition in System 1.
As shown in Table~\ref{tab:case_study}, System 1 makes consecutive errors in two steps due to a lack of deep understanding of the problem and logical analysis.
In System 2, our Knowledge Generator correctly outputs the steps of subtracting first and then adding, allowing Deep Reasoner to obtain the correct answer based on this.
Similarly, System 1 can also reduce errors in System 2 caused by incorrect associations of knowledge or wrong application of it.
Due to limited space, further details are not elaborated.

% Table generated by Excel2LaTeX from sheet 'case'
\begin{table}[tbp]
  \centering
  \setlength{\tabcolsep}{6pt}
  \resizebox{\linewidth}{!}{
    \begin{tabular}{c|p{25em}}
    \toprule
    Q & Last Friday, 13 of the 82 teachers at Rydell Elementary School were sick. There were 9 substitute teachers called in to help. How many teachers were at school that day? \\
    \midrule
    GT & There were 82–13=69 regular teachers at school. If we add the substitute teachers, we find there were 69+9=78 teachers at school that day. \\
    \midrule
    S-1 & There were 13 sick teachers and 9 substitute teachers, so there were a total of \textcolor{red}{13+9=22} teachers not available. Out of the 82 total teachers, 22 were not available, so there were \textcolor{red}{82-22=60} teachers at school that day. \textcolor{red}{\XSolidBrush} \\
    \midrule
    S-2 & Subtract the number of incomplete items from the total to find the complete items. Add the number of additional items to the remaining items to find the total.\newline{}\newline{}There were 82-13=69 teachers at Rydell Elementary School that day. In addition to the 69 teachers, there were 9 substitute teachers. So, the total number of teachers at school that day was 69+9=78. \textcolor{green}{\Checkmark} \\
    \bottomrule
    \end{tabular}%
    }
    \caption{Case study of \OURMODEL{} on the GSM8K dataset (Q: Question, GT: Ground Truth).}
  \label{tab:case_study}%
\end{table}%

\section{Conclusion}
In this work, we propose a novel method \OURMODEL{} with multi-LoRA interaction, which improves the mathematical reasoning performance of student language models like human beings teaching and learning pattern.
\OURMODEL{} explicitly extracts the knowledge of teacher models in the data augmentation stage, and fully utilizes the consistency of System 1 and System 2 in the model distillation stage.
Experimental results show that \OURMODEL{} outperforms the state-of-the-art methods and can be effectively integrated into any CoT distillation model.
In the future, we will explore the following directions:
(1) We will apply the idea of interaction between knowledge and reasoning during the training phase to reduce the inference overhead of models, such as introducing reinforcement learning~\cite{dpo}.
(2) We will use external tools (e.g., compilers) in our approach so that the knowledge generator, reasoning generator, and code generator can verify each other and reduce computational errors to a certain degree.

\appendix

\section*{Acknowledgments}
We thank the reviewers for their insightful comments. This work was supported by National Science Foundation of China (Grant Nos.62376057). All opinions are of the authors and do not reflect the view of sponsors.

\bibliographystyle{named}
\bibliography{ijcai25}

\end{document}